# Practical Machine Learning for Aphasic Discourse Analysis

Jason M Pittman[1], Yesenia Medina-Santos[2], Anton Phillips Jr.[2], Brielle C. Stark[2]
[1] University of Maryland Global Campus
[2] Indiana University Bloomington, Department of Speech, Language and Hearing Sciences

**ABSTRACT**

Analyzing spoken discourse is a valid means of quantifying language ability in persons with aphasia. There are many ways to quantify discourse, one common way being to evaluate the informativeness of the discourse. That is, given the total number of words produced, how many of those are context-relevant and accurate. This type of analysis is called Correct Information Unit (CIU) analysis and is one of the most prevalent discourse analyses used by speech-language pathologists (SLPs). Despite this, CIU analysis in the clinic remains limited due to the manual labor needed by SLPs to code and analyze collected speech. Recent advances in machine learning (ML) seek to augment such labor by automating modeling of propositional, macrostructural, pragmatic, and multimodal dimensions of discourse. To that end, this study evaluated five ML models for reliable identification of Correct Information Units (CIUs, Nicholas & Brookshire, 1993), during a picture description task. The five supervised ML models were trained using randomly selected human-coded transcripts and accompanying words and CIUs from persons with aphasia. The baseline model training produced a high accuracy across transcripts for word vs non-word, with all models achieving near perfect performance (0.995) with high AUC range (0.914 min, 0.995 max). In contrast, CIU vs non-CIU showed a greater variability, with the k-nearest neighbor (k-NN) model the highest accuracy (0.824) and second highest AUC (0.787). These findings indicate that while the supervised ML models can distinguish word from not word, identifying CIUs is challenging. While the ML models exhibited an effective capability, such is may be limited because the models cannot capture *context* or *nuance* within discourse. Continued refinement of these systems may support more efficient and scalable assessment tools for aphasia rehabilitation, reducing the clinical burden while preserving the sensitivity of the diagnostics.

**Introduction**
Discourse, the use of language in connected, meaningful units such as narratives, explanations, and conversations (Armstrong, 2000), is an ecologically valid measure of communicative ability. Because communication in daily life rarely occurs in isolated words or sentences, discourse provides a direct window into the systems that support participation and quality of life. Further, discourse analysis in aphasia[1] offers clinically relevant insight into how language impairments manifest in natural communication, which informs diagnosis (especially for sensitively identifying mildest aphasia), goal setting, and treatment outcomes more effectively than standardized batteries alone (Fromm et al., 2017, Stark, 2019). Yet, despite its clinical importance, discourse analysis remains underused by speech therapists in aphasia rehabilitation due to time, training, and transcription demands (Bryant et al., 2016; Cruice et al., 2020; Stark et al., 2021).

---

[1] Aphasia is a language disorder common after dominant left hemisphere stroke.

Discourse is a multidimensional construct encompassing linguistic, propositional, macrostructural, pragmatic, and multimodal components (Dipper et al., 2021; Stark & Dalton, 2024). Together, these layers capture how individuals produce meaningful, coherent, and socially appropriate language in real-world contexts. Gold-standard analyses of discourse integrate all these levels to yield a comprehensive view of communicative competence (Dipper et al., 2021; Stark & Dalton, 2024). In more mild presentations of aphasia, linguistic skills such as word retrieval and syntax may appear relatively intact, while higher-order discourse processes—e.g., organization of ideas (macrostructural), informativeness (propositional), coherence (macrostructural), perspective-taking (pragmatic), and gesture use (multimodal)—remain impaired. These subtle breakdowns often shape how effective and effortful everyday communication becomes for persons with aphasia (PWA).

Currently, quantifying these higher-order discourse features relies almost entirely on manual coding approaches such as Correct Information Units (Nicholas & Brookshire, 1993), Main Concept Analysis (Nicholas & Brookshire, 1995), and Core Lexicon Analysis (Dalton et al., 2020). While these hand-scored measures are psychometrically robust (Stark et al., 2023, Dalton & Richardson, 2015), they are labor-intensive and impractical for routine clinical use (Bryant et al., 2017; Stark et al., 2021). In contrast, linguistic-level features are now readily extractable through natural language processing (NLP) tools such as CLAN (MacWhinney, 2000) and OpenBrainAI (Themistocleous, 2024), enabling large-scale characterization of aphasia phenotypes (Stark, 2019; Stark & Fukuyama, 2021).

Recent advances in machine learning (ML) extend these capabilities, allowing automated modeling of propositional, macrostructural, pragmatic, and multimodal dimensions of discourse. Such tools can capture the richness of real-world communication more efficiently and reproducibly than manual methods, supporting scalable assessment of communicative functioning in aphasia.

Looking at the present and future, ML has a variety of applications demonstrated in aphasia research. To wit, work by Azevedo et al. (2024) identified four general categories: diagnostic, therapeutic, lexical modeling, and identification of paraphasic errors. Per the authors, the majority (17 of 28) of the literature attempts to apply ML to diagnostic processes (2024). Notably missing from the Azevedo et al. (2024) review, and the literature in general, is the use of ML, specifically large language models (LLMs), in higher-level aphasic discourse analysis.

Again, the need for manual intervention in analyzing clinical sessions for propositional, macrostructural, pragmatic, and multimodal information is a material challenge. Accordingly, the purpose of this work is to construct and validate a practical ML-based capability to reliably identify one well-known and validated propositional metric, informativeness, via the Correct Information Unit (CIU) technique (Nicholas & Brookshire, 1993).

**Methods**
According to Nicholas and Brookshire (1993), coding CIUs requires transcribing a connected speech sample verbatim, counting all intelligible words, and then identifying those that are accurate, relevant, and informative about the topic as CIUs. The process also requires excluding fillers (e.g., "um"), repetitions, and tangential remarks. From these counts, one can calculate metrics such as percent CIUs and CIUs per minute to capture communicative informativeness and efficiency, respectively. CIUs are a cornerstone of aphasia and speech-language pathology research because they provide a standardized, objective index of how effectively individuals convey meaningful content in spontaneous language, independent of grammaticality or fluency (e.g., Stark et al., 2023). This makes CIUs especially sensitive to real-world communicative ability and valuable for tracking change, evaluating treatment outcomes, and comparing discourse performance across individuals and contexts. However, CIU coding remains a time-intensive manual process requiring trained coders. We therefore sought to answer a single research question: *to what extent can traditional ML classification models accurately identify WORD versus NON-WORD and CIU versus NON-CIU?*

To find an answer to this question, we selected five classifier ML models for training and evaluation. The models were Support Vector Machine (SVM) linear, SVM RBF, Decision Tree (DT), k-Nearest Neighbor (k-NN), and Random Forest (RF). All selected models are well understood classification systems and consume the same token-level features (character n-grams, local context, and simple linguistic cues). Yet, the models differ in how each forms decision boundaries.

For instance, a linear SVM is like drawing a straight line to separate two groups as cleanly as possible (Cortes & Vapnik, 1995). An RBF-kernel SVM is similar to a linear SVM except the model allows the line to curve which helps when the groups are mixed in more complex ways (Vapnik, 1998; Schölkopf & Smola, 2002). According to Breiman et al. (1984) and Quinlan (1993) DT models operate like a flowchart of yes or no questions (e.g., "does the token look like a filler?"). Relatedly, a random forest builds many such trees and averages them (Breiman, 2001). Lastly, the k-nearest neighbor (k-NN) model reaches classification decisions by looking at the most similar past examples and reaching a vote on the likely outcome (Cover & Hart, 1967). Overall, because all five models see the same inputs, any differences in results reflect their built-in "habits" for making decisions rather than different data going in.

Keeping this in mind, we trained each model using randomly selected single picture description transcripts (the Cat Rescue, part of the traditional AphasiaBank protocol [MacWhinney et al., 2011]) from Stark's NEURAL Research Lab (Stark et al., 2023; Stark et al., 2025). These transcripts are freely available on AphasiaBank (aphasia.talkbank.org). The transcripts contained no PII, PHI, or information otherwise capable of identifying any person associated with the clinical session. In total, there were 130 discrete Cat Rescue descriptions selected consisting of a mixture of mild, moderate, severe, and very severe aphasias, as well as cognitively healthy individuals.

All models were trained and evaluated in a dedicated Google Colab notebook instance for portability and accessibility. All source code is available in a public repository (*aphasia_classifiers_trainer.ipynb*, Pittman, 2025).

*Procedure*

To affect the purpose of the study, we adhered to a standard procedure of preparing training data (i.e., human-elicited Cat Rescue description transcripts) and then training the five classifier models. Throughout we captured baseline performance metrics.

The transcripts were all coded in the Codes for Human Analysis of Language (CHAT) approach (MacWhinney, 2000), which includes specific markers that aid in computerized analysis using the complementary program CLAN. For our purposes, these additional codes were not needed. Thus, the first step involved data cleaning. Here, we isolated transcript text from participants (i.e., "PAR" marked descriptions whereas "INV" marked investigator speech) and removed any transcription of linguistic features (e.g., fillers, "xxx", &=laughs, punctuation/length cues).

Next, we tokenized each transcript using the standard spaCy package (Honnibal et al., 2020) for Python. The tokenized transcripts were then written to a plaintext file. After data cleaning and tokenization, two trained Research Assistants (authors AP and YMS) labeled the data along two dimensions: WORD or NOT WORD, and if WORD, CIU or NOT CIU. The intent was to leverage WORD and CIU states as features for classification.

As an example, we can take the original utterance,

*uh bog uh dog chased a xxx up a tree*

which was cleaned and tokenized into,

*[ 'bog', 'dog', 'chased', 'a', 'up', 'a', 'tree']*

and then labeled as

*['bog'[WORD][NOT CIU], 'dog'[WORD][CIU], 'chased'[WORD][CIU], 'a'[WORD][NOT CIU], 'up'[WORD][CIU], 'a'[WORD][CIU], 'tree'[WORD][CIU]*

The final step was to upload the tokenized, labeled transcript samples to Google Drive, execute the Python notebooks, and then compile the preliminary results.

**Preliminary Results**

We first evaluated baseline token-level classification for WORD vs. NOT WORD. All five classifiers achieved near-ceiling performance (Accuracy/F1 ≈ 0.995-0.998; Table 1). This convergence suggests a highly separable signal for lexicality in our feature space (character n-grams, local context, and lightweight linguistic markers). Although rank ordering on AUC indicates minor differences in separability, the task-level metrics are effectively saturated. In this context, SVM-rbf and k-NN lead (AUC 0.997 and 0.995), followed by RF (0.989), with linear

SVM and DT lower (0.917 and 0.914). In practice, this implies subsequent gains on WORD are unlikely to be meaningful without harder negatives or domain shift. We therefore treat WORD primarily as a prerequisite gate for CIU (i.e., enforcing CIU=0 when WORD=0).

Table 1.
*Performance metrics for classification of Word and Non-Word*

| Model | Accuracy | Precision | Recall | F1 | AUC |
|---|---|---|---|---|---|
| k-NN | 0.995 | 0.995 | 1.000 | 0.998 | 0.995 |
| SVM-rbf | 0.995 | 0.995 | 1.000 | 0.998 | 0.997 |
| RF | 0.995 | 0.995 | 1.000 | 0.998 | 0.989 |
| SVM-linear | 0.995 | 0.995 | 1.000 | 0.998 | 0.917 |
| DT | 0.995 | 0.995 | 1.000 | 0.998 | 0.914 |

We next assessed CIU vs. NON-CIU, a more challenging, context-dependent decision (Table 2). Here, performance drops to a more informative range, revealing clear model differences. The best F1 is obtained by k-NN (F1=0.889), with SVM-rbf close behind (F1=0.877). RF and SVM-linear form a middle tier (F1=0.867 and 0.865) while DT lags behind the other models (F1=0.796). Notably, SVM-rbf yields the highest AUC (0.797) despite slightly lower F1 than k-NN (AUC 0.787). We take such to indicate stronger ranking quality and, at the same time, suggestive of threshold tuning potentially further narrowing the gap between models. Meanwhile, precision-recall trade-offs are modest at the operating point (e.g., k-NN Prec/Rec 0.886/0.891).

Table 2.
*Performance metrics for classification of CIU and Non-CIU*

| Model | Accuracy | Precision | Recall | F1 | AUC |
|---|---|---|---|---|---|
| k-NN | 0.824 | 0.886 | 0.891 | 0.889 | 0.787 |
| SVM-rbf | 0.806 | 0.875 | 0.880 | 0.877 | 0.797 |
| RF | 0.793 | 0.877 | 0.857 | 0.867 | 0.766 |
| SVM-linear | 0.788 | 0.868 | 0.863 | 0.865 | 0.746 |
| DT | 0.698 | 0.851 | 0.749 | 0.796 | 0.628 |

Taken together, these results show (a) CIU classification is substantially harder than WORD, as expected for informativeness tied to picture-description context and (b) nonlinear decision boundaries (k-NN local neighborhoods; SVM-rbf kernels) provide measurable advantages over linear or single-tree models.

These baselines establish feasibility and identify promising inductive biases for CIU. However, they do not attribute improvements to specific input components. To that end, we conduct a controlled ablation that removes token character n-grams, context character n-grams, handcrafted linguistic markers, and varies the context window (±0, ±1, ±2). We retrain end-to-end under each setting. In the ablation, we pair all evaluations on the identical test items and quantify differences with grouped bootstrap CIs for ΔF1, McNemar's test for accuracy changes, and DeLong for AUC, with Holm-Bonferroni correction across comparisons. This design isolates which features materially contribute to CIU performance beyond the strong WORD gate and clarifies where marginal gains remain achievable.

*Ablation Summary*

To quantify the relative contribution of individual input and contextual features, we conducted structured ablation testing (*aphasia_classifiers_ablation.ipynb*, Pittman, 2025) across five feature configurations and one context-expansion variant. Each classifier was retrained under identical hyperparameters while systematically disabling token-level character features (-token_char), contextual character encodings (-context_char), hand-crafted linguistic features (-handcrafted), and context windowing (-context_window), as well as expanding the local context to two tokens (+ctx2). Performance was evaluated on both WORD vs. NON-WORD as well as CIU vs. NON-CIU tasks. Baseline configurations included all features with a one-token context window.

Table 3.
*CIU Classification Ablation - F1 / AUC Across Models and Configurations*

|  | **Baseline** | **-Token Char** | **-Context Char** | **-Handcrafted** | **-Context Window** | **+Ctx 2** |
|---|---|---|---|---|---|---|
| k-NN | 0.889 / 0.787 | 0.851 / 0.757 | 0.864 / 0.766 | 0.874 / 0.779 | 0.864 / 0.766 | 0.902 / 0.775 |
| SVM rbf | 0.877 / 0.797 | 0.886 / 0.775 | 0.855 / 0.749 | 0.864 / 0.798 | 0.855 / 0.749 | 0.875 / 0.788 |
| RF | 0.867 / 0.766 | 0.849 / 0.754 | 0.875 / 0.753 | 0.845 / 0.755 | 0.875 / 0.753 | 0.819 / 0.731 |
| SVM linear | 0.865 / 0.746 | 0.855 / 0.727 | 0.854 / 0.736 | 0.859 / 0.749 | 0.854 / 0.736 | 0.847 / 0.730 |
| DT | 0.796 / 0.628 | 0.767 / 0.632 | 0.855 / 0.685 | 0.835 / 0.698 | 0.855 / 0.685 | 0.790 / 0.646 |

Table 3 presents the results of our ablation analysis for the CIU classification task. Each configuration isolates the contribution of specific input features from the set of (a) token-level character embeddings; (b) context-based character features; (c) hand-crafted linguistic indicators; and (d) context window size. The F1 score reflects the harmonic mean of precision and recall, while AUC quantifies overall discriminative capacity. Across all models, removing token-level or hand-crafted features produced the largest declines in F1 and AUC, confirming their importance for capturing lexical and semantic informativeness cues. Context-related ablations (-context_char, -context_window) yielded smaller changes, suggesting that fine-grained lexical representation contributes more to CIU detection than additional context. The k-NN model showed a modest improvement with a two-token window (+Ctx 2), indicating that limited local context can enhance similarity-based methods.

Together, these complementary results demonstrate that while lexical discrimination remains trivial for all models, the identification of communicatively informative units depends on richer feature representations that encode both linguistic structure and token-level variability.

Furthermore, we can visualize (Figure 1) the pattern of CIU classification performance across ablation configurations, complementing the numerical results in Table 1. The figure highlights the consistent high performance of the SVM-rbf and k-NN models across settings and makes clear that removing token-level or handcrafted features yields the sharpest drops in F1. This visual summary reinforces that performance differences are systematic rather than incidental and underscores the stability of overall trends across diverse classifier architectures.

**Figure 1**
CIU Classification Performance Across Ablation Configurations

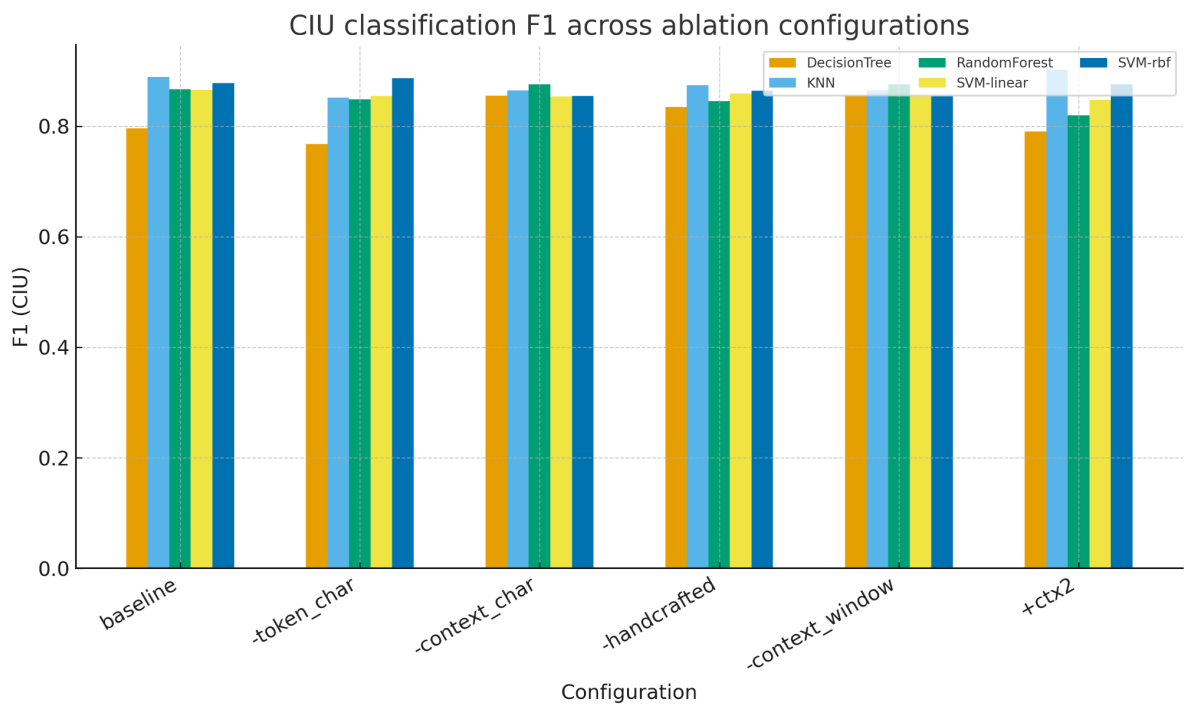

Then, Table 4 summarizes performance on the WORD vs. NON-WORD discrimination task. The task served as a control condition for lexical separability. Accuracy and AUC are reported for each classifier under identical ablation settings. All models achieved near-perfect results (≥ 0.99), and most configurations, including those removing feature groups, retained ceiling-level performance. These findings indicate that distinguishing lexical tokens from non-lexical artifacts is a linearly separable, low-complexity problem requiring minimal contextual or engineered features. Consequently, the observed performance variations in the CIU task (Table 3) reflect sensitivity to semantic and pragmatic information rather than generic degradation in classification capability.

Table 4

*WORD vs. NON-WORD Classification - Accuracy / AUC Across Configurations*

|  | Baseline | -Token Char | -Context Char | -Handcrafted | -Context Window | +Ctx 2 |
|---|---|---|---|---|---|---|
| k-NN | 0.995 / 0.914 | 0.995 / 0.913 | 0.995 / 1.000 | 0.977 / 0.900 | 0.995 / 1.000 | 0.995 / 0.911 |
| SVM rbf | 0.995 / 0.997 | 0.986 / 0.995 | 1.000 / 1.000 | 0.995 / 0.887 | 1.000 / 1.000 | 0.991 / 0.998 |
| RF | 0.995 / 0.989 | 0.991 / 0.986 | 0.995 / 1.000 | 0.982 / 0.902 | 0.995 / 1.000 | 0.986 / 0.988 |
| SVM linear | 0.995 / 0.917 | 0.982 / 0.982 | 1.000 / 1.000 | 0.995 / 0.884 | 1.000 / 1.000 | 0.995 / 0.997 |
| DT | 0.995 / 0.915 | 0.991 / 0.914 | 1.000 / 1.000 | 0.968 / 0.903 | 1.000 / 1.000 | 0.986 / 0.912 |

Figure 2 demonstrates the relative impact on F1 from the baseline configuration for each model, summarizing how sensitive each classifier is to feature removal or context expansion. The visualization succinctly captures the magnitude and direction of these effects, showing that token-level and handcrafted ablations consistently reduce F1 while context-related changes remain close to zero. By expressing results in relative terms, the figure clarifies that observed performance losses stem from information-bearing feature reductions rather than random variation.

Taken together, the ablation results demonstrate that model success in distinguishing communicatively informative language depends on representational fidelity rather than architectural complexity. All classifiers performed equivalently on the shallow lexical task, but only those retaining token-level and linguistically grounded features maintained strong CIU discrimination. These findings suggest that informativeness in aphasic discourse is encoded primarily through fine-grained lexical and morphological cues that capture word accuracy and contextual relevance. Broader context windows and additional model depth contributed little once these features were present, indicating that improvements in informativeness modeling should prioritize richer linguistic feature design and localized representation learning over wider contextual expansion.

**Figure 2**
Impact of Feature Ablations on CIU Classification Performance

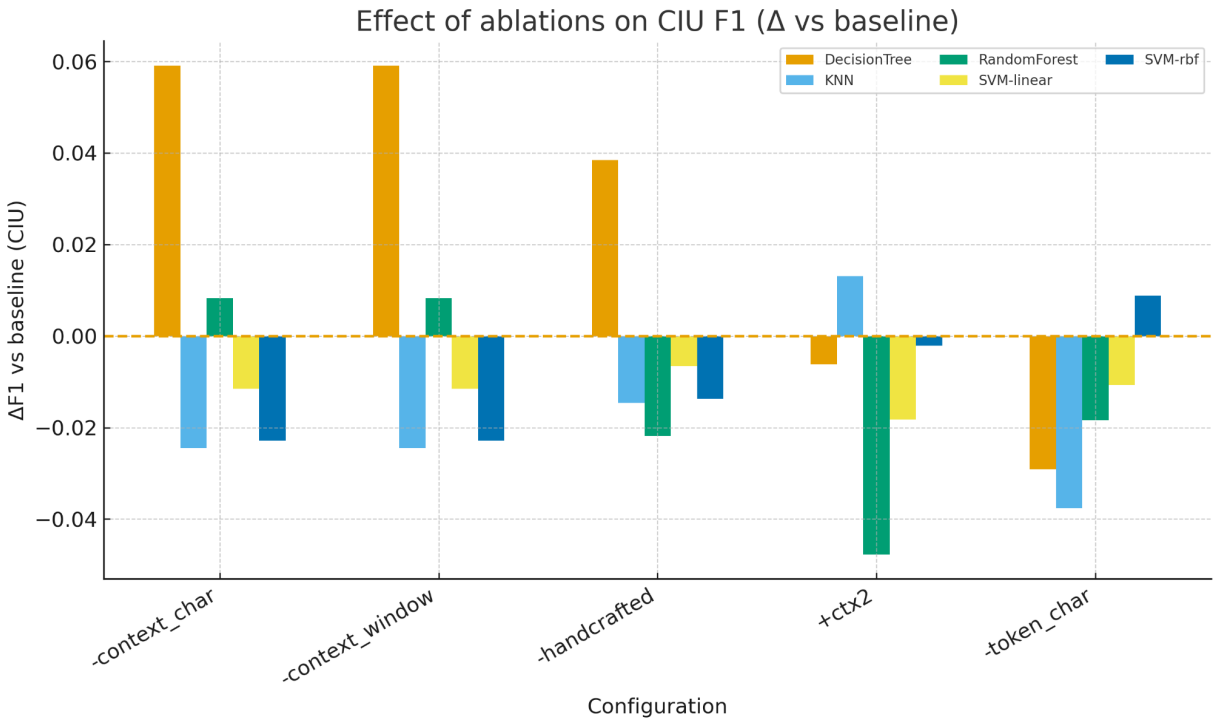

In summary, the ML models demonstrated ceiling-level performance for basic lexical discrimination (word vs. non-word), confirming this component of discourse analysis can be (and has been in many NLP tools) automated with high confidence. However, distinguishing informative from non-informative words, i.e., CIUs from non-CIUs, a more context-sensitive and cognitively loaded judgment, proved substantially more difficult. Even the best-performing models (k-NN and SVM-rbf) achieved only moderate-to-high accuracy, indicating that informativeness in discourse depends on linguistic and semantic precision beyond surface lexical cues. Meanwhile, feature ablation analyses revealed that token-level and linguistically engineered features contribute most strongly to CIU detection, whereas expanding contextual windows or increasing model complexity yielded minimal benefits. Overall, these findings suggest that while traditional ML approaches can reliably support low-level segmentation tasks, successful automation of discourse informativeness will require models that encode richer linguistic structure and pragmatic context, underscoring both the promise and current limitations of ML-assisted discourse analysis in aphasia.

**Conclusions**
Discourse analysis in aphasia is a method to investigate how, and to what extent, language impairments manifest in natural communication. One task employed by the method involves asking a patient to tell a story about elements and events displayed in a standardized picture. Responses are recorded, transcribed, and then evaluated by a trained specialist. While diagnostically robust, existing analytical techniques rely on manual coding approaches. Here, a problem exists insofar as the techniques are labor-intensive and impractical for routine clinical use (Bryant et al., 2017; Stark et al., 2021). Meanwhile, the rise of AI in healthcare has brought about great interest in clinical applications of ML in aphasia. Researchers in this domain have demonstrated ML-based solutions for diagnostics, therapeutics, lexical modeling, and identification of paraphasic errors (Azevedo et al., 2024).

Considering the above setting, we set out to construct and validate a practical ML-based capability to reliably identify CIU in Cat Rescue description text transcripts. Towards this aim, we collected a sample of 130 Cat Rescue descriptions from prior work (Stark et al., 2023) and trained five ML models: SVM-linear, SVM-rbf, DT, k-NN, and RF. Our results demonstrate baseline model performance showing two critical contrasts in task complexity. The near-perfect results across all classifiers on the WORD vs. NON-WORD discrimination task demonstrated that lexical segmentation in aphasic transcripts is a trivial, linearly separable problem requiring minimal contextual or engineered features. In contrast, the CIU vs. NON-CIU classification task revealed substantially lower scores with F1 values in the 0.79-0.89 range. We suggest this indicates communicative informativeness depends on more nuanced, higher-order linguistic signals.

Subsequent ablation testing confirmed that this difficulty arises not from model instability but from the inherently multi-level nature of the CIU construct. When token-level and handcrafted linguistic features were removed, performance declined markedly across all models, underscoring the importance of lexical precision and structured linguistic cues in modeling informativeness. Context-related ablations and wider context windows produced negligible or inconsistent effects, suggesting that discourse-level context beyond a single token contributes little additional discriminative power.

Collectively, these findings indicate that informative language in aphasia is locally encoded and harbors lexical and morphological structure rather than distributed broadly across context. From a methodological perspective, the findings validate that traditional classifiers can achieve high reliability in this domain when equipped with appropriately engineered features, while further gains in CIU detection will likely come from refining linguistic representations rather than expanding context or increasing model complexity.

*Limitations*
Our classifiers were trained and evaluated on a narrow corpus of picture-description transcripts with a small number of speakers and tokens. As a result, performance may not generalize to other discourse tasks (e.g., narratives, conversations), different clinical populations, dialects, or recording conditions. The goal of this work is to establish methodological viability rather than broad generalization, but external validation on additional corpora remains necessary.

Further, we intentionally prioritized a controlled baseline over exhaustive optimization. Thus, we did not conduct post-hoc, error-driven feature engineering, extensive hyperparameter searches, probability calibration, or operating-point tuning (e.g., threshold selection for CIU). Consequently, underperforming models could likely be improved; however, our intent here is to compare inductive biases under matched inputs rather than to maximize absolute scores.

The classic ML models used are only weakly context-aware (short ±1-2 token windows) and operate over text alone. They do not encode discourse structure (e.g., topic flow, coreference resolution), task grounding (the stimulus picture), prosody, or acoustic cues, all of which can influence CIU judgments. Thus, the models can decide *that* a token is a WORD or CIU but have limited access to *why* in the pragmatic context.

Finally, CIU labeling is inherently subjective. Whilst raters in the Stark lab have demonstrated high reliability for coding these CIUs (see Stark et al., 2023), small differences between raters could lead to WORD/CIU class imbalances in the training data causing some metrics (e.g., AUC) to be misleading. Here, precision-recall curves and calibration analysis would strengthen interpretation.

### Recommendations

We offer the following four recommendations given the preliminary results and limitations. The initial two recommendations aim to improve model performance based on the current architecture. Then, we offer two recommendations for pilot implementations of performant models in clinical settings.

Foremost, the ablations reveal small windows carrying the majority of the classic-model signal. Consequently, we recommend configuring such in the ±1-2 tokens band for lightweight deployments to preserve short-range context. Further, we recommend enforcing CIU as zero whenever WORD is zero as doing so may consistently improve CIU precision without retraining.

In the context of pilot implementations, we recommend normalizing fillers, 'xxx', and laughter tokens during preprocessing. Lastly, we foresee a need to retain a human-in-the-loop for low confidence outcomes. Thus, we recommend pilot testing include a calibrated *Confidence* score so that, when near a configurable threshold (e.g., 0.4 - 0.6), the decision of whether a token is a WORD and CIU is routed to a clinician.

Overall, the preliminary results are to be understood only within the confines of the stated purpose. In other words, this work reveals what may be an avenue for practical application of ML to clinical assessment of people with aphasia. We imagine a variety of future work as possible to extend our preliminary results as well as address the stated limitations.

### Clinical Implications

This study demonstrates the early feasibility and current limits of using classifier ML tools to automate the identification of CIUs in aphasic discourse. For clinicians, these findings underscore that while ML can already replicate basic linguistic discrimination tasks (e.g., word vs. non-word), it still struggles to capture the contextual and pragmatic nuances that define informativeness in connected speech. The clinical takeaway is that such automated tools are not yet replacements for clinician judgment but could soon function as assistive technologies to accelerate the transcription and scoring workflow. For example, an ML-assisted pipeline could pre-label CIUs with confidence scores, allowing SLPs to verify rather than manually code every token, thereby reducing analysis time while maintaining diagnostic rigor. By embedding these models into existing clinical software or research databases, SLPs could increase efficiency in discourse-based outcome measurement and scale assessment to larger caseloads or multi-site studies. However, the performance variability across models highlights the ongoing need for domain-specific feature engineering and clinician oversight, especially in capturing subtle communicative informativeness. In practice, ML-assisted discourse analysis may soon offer a hybrid *human-in-the-loop* model, where algorithms handle low-level linguistic segmentation while clinicians focus on interpretation, goal-setting, and treatment planning. As such systems mature, they hold promise for democratizing access to discourse-based metrics, enabling faster and more ecologically valid assessments that preserve the clinical insight of expert human analysis.

*Future Work*
Indeed, future work is necessary to test the trained models on novel transcripts not in the original train-test dataset, as well as extrapolate beyond the picture description task. Therefore, we offer four ideas for additional research intended to address limitations directly, extend the method, or to construct alternative methods addressing the same problem.

Certainly k-NN and SVM-rbf show performance metrics that warrant additional study. Work can be done not only on feature engineering and tuning but exposing at least those two models to novel data may reveal new findings. This includes evaluation on additional aphasia corpora, discourse tasks (narratives/dialogue), dialects, and recording conditions.

As well, this work could be replicated with the addition of quantified model annotator disagreement explicitly (soft labels) and performance capture for contentious tokens. Such might include performance analysis by token type (e.g., fillers, proper nouns, partial words), speaker characteristics, and aphasia severity to foreground systematic gaps in the method.

While some traditional classification models demonstrate sufficient performance metrics to warrant continuing this work, there are potential alternative solutions as well. Additional future work should be done to validate the applicability of large language models (LLMs) to CIU identification. There are two potential paths here. On one hand, one could *instruct* the LLMs on the standardized CIU analysis procedure from Nicholas and Brookshire (1993). On the other hand, one could systematically expose the LLMs to previously analyzed discourse transcripts and use *concept induction* to discover the procedure. Likewise, neurosymbolic approaches combined with supervised learning (Dong & Sifa, 2023) and knowledge graphs (DeLong et al., 2024) offer potential avenues for high accuracy, high reliability analysis of aphasic discourse.

**Acknowledgments**
We are grateful for the careful editing and proofreading assistance provided by Emilia Aram and Hilary Yang.